\documentclass{article}

\PassOptionsToPackage{numbers,compress}{natbib}

\usepackage[preprint]{neurips_2026}

\usepackage[utf8]{inputenc}
\usepackage[T1]{fontenc}
\usepackage{url}
\usepackage{booktabs}
\usepackage{amsfonts}
\usepackage{nicefrac}
\usepackage{microtype}

\usepackage{float}
\usepackage{pifont}
\usepackage{footnote}
\usepackage{enumitem}
\usepackage{bm}
\usepackage{arydshln}
\usepackage{multicol}
\usepackage{multirow}
\usepackage{color}
\usepackage{xcolor}
\usepackage{colortbl}
\usepackage{soul}
\usepackage{bbding}
\usepackage{makecell}
\usepackage{mathtools}
\usepackage{amssymb}
\usepackage{graphicx}
\usepackage{amsmath}
\usepackage{threeparttable}
\usepackage{algorithm}
\usepackage{algorithmic}

\usepackage[most]{tcolorbox}

\definecolor{stepblue}{HTML}{C7547E}       
\definecolor{stepbluebg}{HTML}{FDF0F5}     
\definecolor{steporange}{HTML}{C47AA0}     
\definecolor{steporangebg}{HTML}{FAF0F6}   
\definecolor{stepgreen}{HTML}{A87AB8}      
\definecolor{stepgreenbg}{HTML}{F5EFF8}    
\definecolor{stepred}{HTML}{8B6AAE}        
\definecolor{stepredbg}{HTML}{F0ECF6}      
\definecolor{steppurple}{HTML}{6A1B9A}
\definecolor{steppurplebg}{HTML}{F3E8F9}
\definecolor{algframe}{HTML}{D4B8CC}       
\definecolor{algtitle}{HTML}{8B5E83}       

\newtcolorbox{algbox}[1]{%
  enhanced,
  colback=white,
  colframe=algframe,
  coltitle=white,
  fonttitle=\bfseries\small,
  title={#1},
  colbacktitle=algtitle,
  boxrule=0.6pt,
  arc=2pt,
  left=3pt, right=3pt, top=2pt, bottom=2pt,
  toptitle=2pt, bottomtitle=2pt,
}

\newenvironment{algstepenv}[3]{%
  \begin{tcolorbox}[enhanced,
    colback=#3, colframe=#2, colbacktitle=#2, coltitle=white,
    fonttitle=\bfseries\footnotesize, title={#1},
    boxrule=0.4pt, arc=1.5pt,
    left=3pt, right=3pt, top=1pt, bottom=1pt,
    toptitle=1.5pt, bottomtitle=1.5pt,
    before skip=2pt, after skip=2pt,
    fontupper=\small]%
}{%
  \end{tcolorbox}%
}

\definecolor{citecolor}{HTML}{0071BC}
\definecolor{linkcolor}{HTML}{ED1C24}
\usepackage[colorlinks,
  anchorcolor=red,
  citecolor=citecolor,
  linkcolor=linkcolor,
]{hyperref}

\definecolor{highlight}{rgb}{0, 255, 0}

\newcommand{\dataset}{{\fontfamily{ppl}\selectfont RadThinking}}

\newcommand{\numofpatient}{9{,}131}
\newcommand{\numofnormal}{2{,}077}
\newcommand{\numofct}{20{,}362}

\newcommand{\numofclass}{19}
\newcommand{\numofhospital}{10}
\newcommand{\nunofyearfollowup}{1}

\newcolumntype{P}[1]{>{\centering\arraybackslash}p{#1}}
\newlength\savewidth

\title{\dataset: A Dataset for Longitudinal Clinical Reasoning in Radiology}

\author{
\bf Wenxuan Li\textsuperscript{1} \quad
\bf Pedro R. A. S. Bassi\textsuperscript{1} \quad
\bf Xinze Zhou\textsuperscript{1} \quad
\bf Jakob Wasserthal\textsuperscript{2} \\
\bf Alan L. Yuille\textsuperscript{1} \quad
\bf Zongwei Zhou\textsuperscript{1,3}\thanks{Correspondence to: Zongwei Zhou (\href{mailto:zzhou82@jh.edu}{\textsc{zzhou82@jh.edu}})} \\[2.5mm]
\textsuperscript{1}Department of Computer Science, Johns Hopkins University \\
\textsuperscript{2}Clinic of Radiology and Nuclear Medicine, University Hospital Basel \\
\textsuperscript{3}Department of Oncology, Johns Hopkins School of Medicine \\[2.0mm]
{\small \texttt{Code, Models \& Data:} \href{https://huggingface.co/datasets/wenxuanchelsea/RadThinking}{\texttt{https://huggingface.co/datasets/wenxuanchelsea/RadThinking}}}
}

\begin{document}

\maketitle

\begin{abstract}

Cancer screening is a reasoning task. A radiologist observes findings, compares them to prior scans, integrates clinical context, and reaches a diagnostic conclusion confirmed by pathology. We present \dataset, a Visual Question Answering (VQA) dataset that makes this reasoning explicit and trainable. \dataset\ releases VQA pairs at three difficulty tiers. \emph{Foundation VQAs} are atomic perception questions. \emph{Single-step reasoning VQAs} apply one clinical rule. \emph{Compositional VQAs} require multi-step chain-of-thought to reach a guideline category such as LI-RADS-5. For every compositional VQA, we release the chain of foundation VQAs that solves it. The chain follows the rules of the governing clinical reporting standard. The dataset spans \numofct\ CT scans from \numofpatient\ patients across 43 cancer groups, plus \numofnormal\ verified healthy controls with $\geq$\nunofyearfollowup-year follow-up. To our knowledge, \dataset\ is the first cancer-screening VQA corpus that stratifies questions by reasoning depth and grounds compositions in clinical reporting standards. The foundation tier supplies atomic perception supervision. The compositional tier supplies chain-of-thought data and verifiable rewards for reinforcement-learning recipes such as DeepSeek-R1 and OpenAI o1. \dataset\ enables systematic training and evaluation of whether AI systems can \emph{reason} about cancer, not merely detect it.

\end{abstract}

\section{Introduction}\label{sec:introduction}

\begin{figure}[t]
  \centering
  \includegraphics[width=0.85\linewidth]{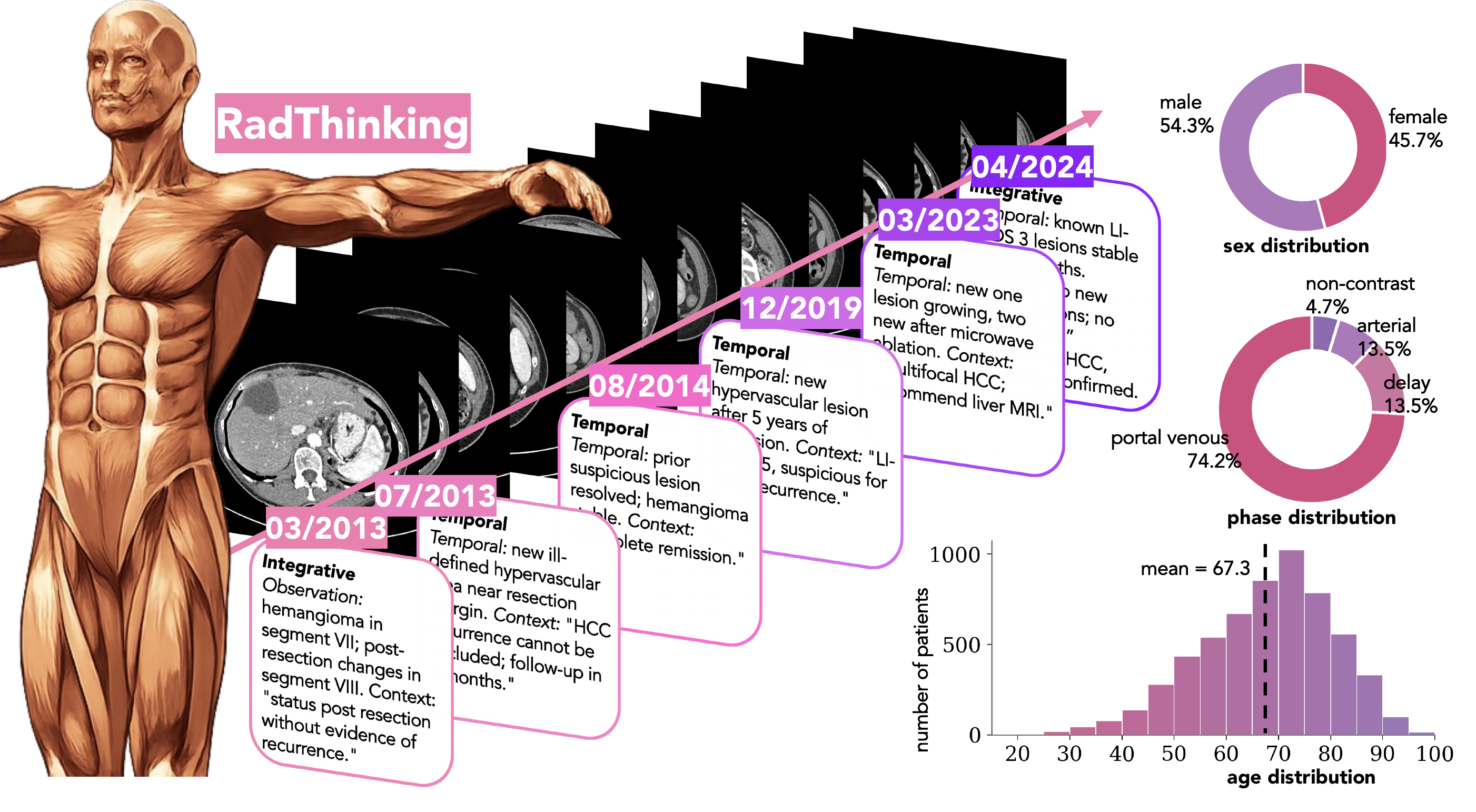}
  \caption{\textbf{Overview of \dataset.} \textit{Left:} the reasoning trajectory of an illustrative hepatocellular carcinoma (HCC) patient. We monitored this patient across 26~CT scans over 11~years (2013--2024). Six selected timepoints show how reasoning complexity evolves: post-resection baseline (Scan~1), a new suspicious lesion (Scan~2), resolution confirming a benign finding (Scan~5), recurrence after 5~years of remission (Scan~17), progression after ablation (Scan~24), and stability that downgrades concern (Scan~26). Each scan carries a four-step reasoning chain: observations, temporal comparison, clinical context, and a pathology-confirmed conclusion. \textit{Right:} dataset characteristics. \dataset\ has \numofct\ CT scans from \numofpatient\ patients across \numofclass\ organ screening targets (43 cancer groups). The distributions cover patient age, sex, and contrast phase.
  }
  \label{fig:dataset_overview}
\end{figure}

A radiologist screening a CT scan for cancer does not simply look for bright or dark spots. They reason. They observe findings, compare to prior scans, integrate clinical context, and reach a diagnosis confirmed by pathology. This chain separates screening from pattern matching.

Public CT datasets reduce this process to a perception task. They provide a scan and a segmentation mask~\cite{bilic2019liver,heller2019kits19,li2025pants,menze2015multimodal,antonelli2021medical}. Evaluation asks one question: did the model find the tumor? They lack longitudinal trajectories. They lack radiology reports. They lack clinical variables. Models trained on them are optimized for perception, not reasoning~\cite{bassi2025learning,liu2024universal,bassi2025scaling,xia2022felix}. The hardest cancer-screening cases need exactly this reasoning. Early-stage tumors are ambiguous on a single scan. Temporal comparison and clinical context are what make detection reliable~\cite{li2026early,li2023early}.

We introduce \dataset\ (Fig.~\ref{fig:dataset_overview}), a VQA dataset for multicancer screening at three difficulty tiers. Hard compositional VQAs decompose into chains of foundation VQAs. The decomposition follows the rules of the governing clinical reporting standard. This mirrors how compositional reasoning in general AI is built from atomic primitives~\cite{press2022selfask,zhou2023leasttomost,khot2023decomposed,gupta2023visprog,suris2023vipergpt}. \dataset\ trains vision-language models (VLMs) along two paths. The foundation tier supplies atomic visual skills~\cite{hong2025atomicvisual}. The compositional tier supplies chain-of-thought data~\cite{wei2022cot,xu2024llavacot,shao2024visualcot} and verifiable rewards for RL recipes such as DeepSeek-R1~\cite{guo2025deepseekr1} and OpenAI o1~\cite{openai2024o1}. The two tiers form a curriculum~\cite{bengio2009curriculum,deng2025currreft}. We release \dataset\ under CC BY-NC-SA 4.0.

\noindent\textbf{Related work.} Table~\ref{tab:related} positions \dataset\ against representative datasets across seven properties. We summarize each group below and identify the gap that \dataset\ fills.

\begin{table}[t]
  \centering
  \caption{\textbf{\dataset\ in context.} Comparison against representative cancer-screening, medical-VQA, and clinical-reasoning resources across seven properties: 3D imaging, voxel-wise tumor masks, paired radiology reports, longitudinal scans, VQA pair release, multi-tier difficulty stratification, and pathology-confirmed ground truth. \ding{51}~present, \ding{55}~absent.}
  \label{tab:related}
  \scriptsize
  \begin{tabular}{p{0.35\textwidth} p{0.10\textwidth} *{7}{c}}
    \toprule
    \textbf{Dataset} & \textbf{Modality} & \textbf{3D} & \textbf{Voxel} & \textbf{Report} & \textbf{Long.} & \textbf{VQA} & \textbf{Tiers} & \textbf{Path.} \\
    \midrule
    \multicolumn{9}{l}{\textit{Cancer segmentation datasets (perception only).}} \\
    KiTS/LiTS/MSD/PanTS~\cite{heller2019kits19,bilic2019liver,antonelli2021medical,li2025pants} & CT & \ding{51} & \ding{51} & \ding{55} & \ding{55} & \ding{55} & \ding{55} & \ding{51} \\
    AbdomenAtlas~2.0~\cite{chen2025scaling} & CT & \ding{51} & \ding{51} & \ding{55} & \ding{55} & \ding{55} & \ding{55} & \ding{55} \\
    ULS23~\cite{de2025uls23} & CT & \ding{51} & \ding{51} & \ding{55} & \ding{55} & \ding{55} & \ding{55} & \ding{55} \\
    \midrule
    \multicolumn{9}{l}{\textit{Imaging paired with text.}} \\
    CT-RATE~\cite{hamamci2026generalist} & Chest CT & \ding{51} & \ding{55} & \ding{51} & \ding{55} & \ding{55} & \ding{55} & \ding{55} \\
    MIMIC-CXR~\cite{johnson2019mimic} & X-ray & \ding{55} & \ding{55} & \ding{51} & \ding{55} & \ding{55} & \ding{55} & \ding{55} \\
    RadGPT~\cite{bassi2025radgpt} & Abd. CT & \ding{51} & \ding{51} & \ding{51} & \ding{55} & \ding{55} & \ding{55} & \ding{55} \\
    \midrule
    \multicolumn{9}{l}{\textit{Medical visual question answering.}} \\
    VQA-RAD/SLAKE/PathVQA~\cite{lau2018dataset,liu2021slake,he2020pathvqa} & 2D mixed & \ding{55} & \ding{55} & \ding{55} & \ding{55} & \ding{51} & \ding{55} & \ding{55} \\
    OmniMedVQA~\cite{hu2024omnimedvqa} & 2D mixed & \ding{55} & \ding{55} & \ding{55} & \ding{55} & \ding{51} & \ding{55} & \ding{55} \\
    M3D-VQA~\cite{bai2024m3d} & 3D Med. & \ding{51} & \ding{51} & \ding{55} & \ding{55} & \ding{51} & \ding{55} & \ding{55} \\
    3D-RAD~\cite{gai20253d} & 3D CT & \ding{51} & \ding{55} & \ding{55} & \ding{51} & \ding{51} & \ding{55} & \ding{55} \\
    DeepTumorVQA~\cite{chen2025vision} & CT & \ding{51} & \ding{55} & \ding{55} & \ding{55} & \ding{51} & \ding{55} & \ding{55} \\
    Kvasir-VQA-x1~\cite{gamage2025kvasirvqax1} & Endoscopy & \ding{55} & \ding{55} & \ding{55} & \ding{55} & \ding{51} & \ding{51} & \ding{55} \\
    \midrule
    \multicolumn{9}{l}{\textit{Reasoning chains for medicine.}} \\
    MedReason / HuatuoGPT-o1~\cite{wu2025medreason,chen2024huatuogpt} & Text & \ding{55} & \ding{55} & \ding{55} & \ding{55} & \ding{51} & \ding{55} & \ding{55} \\
    PhysicianBench~\cite{liu2026physicianbench} & EHR text & \ding{55} & \ding{55} & \ding{55} & \ding{51} & \ding{55} & \ding{55} & \ding{55} \\
    CheXthought~\cite{sharma2026chexthought} & X-ray & \ding{55} & \ding{55} & \ding{51} & \ding{55} & \ding{55} & \ding{55} & \ding{55} \\
    \midrule
    \rowcolor{stepbluebg}\textbf{\dataset\ (ours)} & \textbf{CT} & \ding{51} & \ding{51} & \ding{51} & \ding{51} & \ding{51} & \ding{51} & \ding{51} \\
    \bottomrule
  \end{tabular}
\end{table}

\emph{Cancer segmentation datasets.} KiTS, LiTS, PanTS, BraTS, MSD~\cite{heller2019kits19,bilic2019liver,li2025pants,menze2015multimodal,antonelli2021medical} pair scans with voxel masks. Multi-organ atlases~\cite{li2024abdomenatlas,li2024well,chen2025scaling,liu2023flare,bassi2024touchstone,de2025uls23} scale up. None pair scans with text or reasoning.

\emph{Imaging plus text.} CT-RATE, MIMIC-CXR, CT2Rep, RadGPT~\cite{hamamci2026generalist,johnson2019mimic,hamamci2024ct2rep,bassi2025radgpt} pair imaging with reports. They cover one body region and lack reasoning structure.

\emph{Medical VQA.} 2D resources~\cite{lau2018dataset,liu2021slake,he2020pathvqa,zhang2023pmcvqa,hu2024omnimedvqa} pair images with short factual answers. 3D extensions~\cite{bai2024m3d,gai20253d,chen2025vision,wang2025medframeqa} reach volumes but do not stratify questions by reasoning depth. Kvasir-VQA-x1~\cite{gamage2025kvasirvqax1} stratifies questions for endoscopy. Compositional-VQA work in general AI~\cite{press2022selfask,zhou2023leasttomost,khot2023decomposed,gupta2023visprog,suris2023vipergpt,andreas2016nmn,selvaraju2020squint,johnson2017clevr,hudson2019gqa} establishes the decomposition primitive. It has not been operationalized for cancer screening.

\emph{Reasoning chains in medicine.} MedReason, HuatuoGPT-o1, Med-PRM~\cite{wu2025medreason,chen2024huatuogpt,yun2025med} are text-only. PhysicianBench~\cite{liu2026physicianbench} decomposes EHR tasks into 670 checkpoints, demonstrating that visible chains expose model failures, but operates without imaging. CheXthought~\cite{sharma2026chexthought} releases free-form CoT for chest X-ray, not structured VQA. Medical VLMs Merlin, RadFM, Med-Gemini~\cite{blankemeier2024merlin,wu2025radfm,saab2024medgemini} establish broader radiology VLM training.

\emph{Innovation.} \dataset\ is the first resource that releases cancer-screening VQA pairs at multiple reasoning-depth tiers, decomposes each compositional question into foundation VQAs organized by clinical reporting standards, and anchors conclusions to pathology with longitudinal voxel-grounded imaging.

\section{The \dataset\ Dataset}\label{sec:dataset}

\subsection{What \dataset\ Contains}\label{sec:contents}

The primary released artifact is a corpus of (CT scan, question, answer) triples at three difficulty tiers (\S\ref{sec:vqa_tiers}). Compositional triples additionally carry a chain of foundation triples that solves them. Supporting artifacts are released alongside: standardized NIfTI CT volumes, voxel-wise tumor masks across \numofclass\ organ screening targets (twelve organs with no prior public CT tumor annotations), paired de-identified radiology reports and clinical variables, pathology labels for cancer-positive patients, and a confirmation of $>$\nunofyearfollowup-year cancer-free follow-up for verified healthy patients.

Table~\ref{tab:schema} summarizes the per-patient JSON. It records the four-step reasoning chain that underlies every compositional VQA. Each trace stores the four steps (\S\ref{sec:reasoning_traces}) plus metadata, parsed report, and risk category. The full schema appears in Appendix~\ref{app:json_schema}. The JSON is the source from which the VQA pairs are generated (\S\ref{sec:training}).

\subsection{Cohort, Annotation, and Multimodal Data}\label{sec:cohort}

\dataset\ contains \numofpatient\ patients and \numofct\ pelvic, abdominal, and thoracic CT scans from \numofhospital\ European institutions, acquired 2012 to 2025 under IRB and ethics approval. The median follow-up is 1.17 years per patient. The cancer-positive cohort has confirmed malignancies across 43 cancer groups spanning \numofclass\ organ screening targets, with all prior scans retained per patient. The verified-healthy cohort has $>$\nunofyearfollowup-year follow-up after the last CT. We split strictly at the patient level, stratified by cancer type, organ target, and institution: $N_{\text{train}} = 7{,}305$ patients (16{,}290 scans) and $N_{\text{test}} = 1{,}826$ patients (4{,}072 scans).

CT volumes are released in standardized NIfTI format with harmonized orientation and voxel spacing. Voxel-wise tumor masks cover all \numofclass\ organ screening targets; twelve of these targets have no prior public CT tumor annotation\footnote{The seven organs with existing public CT tumor masks are liver~\cite{bilic2019liver}, kidney~\cite{heller2019kits19}, and pancreas, colon, lung, spleen, prostate~\cite{antonelli2021medical}. The twelve targets without prior public masks are thyroid, breast, esophagus, gallbladder, stomach, duodenum, adrenal, bladder, uterus, ovary, lymph node, and bone.}. Annotation used a three-stage protocol: 28 radiologist residents produced initial masks via MONAI Label~\cite{diaz2024monai}, two of eight board-certified radiologists independently reviewed each case, and a separate radiologist adjudicated discrepancies. The protocol builds on prior scalable CT annotation~\cite{qu2023annotating,li2025scalemai,zhang2024leveraging,chou2024acquiring,bassi2025label,chen2026large}. Mean inter-reviewer Dice on a 200-patient validation cohort is 62.2\% (per-organ in Appendix~\ref{app:pipeline}). Cases with Dice $<$ 0.30 receive an \emph{ambiguity flag} that feeds the complexity stratification (\S\ref{app:complexity}).

\begin{table}[t]
  \centering
  \caption{\textbf{The per-patient JSON file in \dataset.} Every patient is a single JSON record. Patient-level fields summarize the patient. The \texttt{reasoning\_traces} list contains one structured reasoning chain per CT scan. Step~1 to Step~4 mirror the chain definition in Eq.~\ref{eq:trace}.}
  \label{tab:schema}
  \scriptsize
  \begin{tabular}{p{0.22\textwidth} p{0.72\textwidth}}
    \toprule
    \textbf{field} & \textbf{description} \\
    \midrule
    \multicolumn{2}{l}{\emph{Patient-level fields}} \\
    \texttt{patient\_id} & anonymized patient identifier; one folder of CT volumes per id \\
    \texttt{primary\_cancer} & resolved cancer type, confidence, source, all candidate scores, metastasis flag \\
    \texttt{clinical\_history} & list of prior diagnoses, procedures, and oncological status \\
    \texttt{num\_scans}, \texttt{date\_range} & length and time span of the longitudinal sequence \\
    \texttt{reasoning\_traces} & list of one reasoning chain per CT scan (fields below) \\
    \midrule
    \multicolumn{2}{l}{\emph{Per-scan trace fields}} \\
    \texttt{metadata} & scan id, accession, date, scan index, age, sex, contrast phase, malignancy/metastasis flags \\
    \texttt{step1\_observations} & list of findings; each finding stores organ, location, size, attenuation, tumor type, certainty, malignancy/metastasis flags, governing clinical standard (\S\ref{app:step1}) \\
    \texttt{step2\_temporal} & per-lesion change labels (\textsc{new}, \textsc{growing}, \textsc{stable}, \textsc{shrinking}, \textsc{resolved}); interval to prior scan; counts of new, matched, and resolved findings (\S\ref{app:step2}) \\
    \texttt{step3\_clinical\_context} & parsed report (findings, impression, recommendation), RECIST assessment, organ-specific risk category, structured clinical variables, raw report text (\S\ref{app:step3}) \\
    \texttt{step4\_conclusion} & primary cancer with confidence, organ-level diagnosis, ICD-10 code, metastatic disease flag (\S\ref{app:step4}) \\
    \texttt{reasoning\_complexity} & one of \textsc{perceptual}, \textsc{temporal}, \textsc{integrative}, \textsc{ambiguous} (\S\ref{app:complexity}) \\
    \bottomrule
  \end{tabular}
\end{table}

Each scan is paired with the de-identified radiology report and the clinical variables available at imaging time. A parsing pipeline (Appendix~\ref{app:pipeline}) extracts findings, impression, and recommendation. Pathology serves as the ground-truth conclusion (\S\ref{app:step4}) for cancer-positive patients; absence of cancer over $>$\nunofyearfollowup-year follow-up is the negative ground truth for healthy patients. All chain inputs reflect only information available at or before imaging time, which prevents future-information leakage.

\section{VQA Tiers and Compositional Structure}\label{sec:vqa_tiers}

\dataset\ organizes its VQA pairs into three difficulty tiers. \emph{Foundation VQAs} (\S\ref{sec:foundation_vqa}) are atomic perception questions whose answers come from a single annotated field. \emph{Single-step reasoning VQAs} (\S\ref{sec:single_step_vqa}) apply one explicit clinical rule to one foundation observation. \emph{Compositional VQAs} (\S\ref{sec:compositional_vqa}) require multiple foundation answers to be composed via the rules of a clinical reporting standard. The three tiers form a curriculum from atomic skills to multi-step clinical reasoning. Box~1 makes the chain visible. It shows a complete reasoning trace for one compositional VQA, decomposed into foundation VQAs grouped by step. This mirrors how recent agentic medical benchmarks expose the reasoning chain through structured checkpoints~\cite{liu2026physicianbench,sharma2026chexthought}, but with vision in the loop and with each checkpoint formulated as a verifiable VQA pair.

\begin{figure}[t]
\begin{algbox}{Box 1: Reasoning Chain Trace for a Compositional VQA (Patient HCC, Scan~17, 2019-12)}
\phantomsection\label{box:trace}%

\smallskip
{\footnotesize\textbf{Compositional VQA.} \emph{What is the LI-RADS category of the new hepatic lesion in segments VI/VII?}\\
\textbf{Patient context.} HCC post-resection (2013); chronic at-risk for HCC; six prior scans.}

\begin{algstepenv}{\ding{192}~Step 1: Imaging Observations (Foundation VQAs)}{stepblue}{stepbluebg}
{\small
\textbf{Q$_1$:} Are any new hepatic lesions present?\quad \textbf{A:} Yes, in segments VI/VII.\\
\textbf{Q$_2$:} What is the lesion size?\quad \textbf{A:} 1.4\,cm.\\
\textbf{Q$_3$:} Is arterial-phase hyperenhancement present?\quad \textbf{A:} Yes.\\
\textbf{Q$_4$:} Is portal-venous washout present?\quad \textbf{A:} Yes.\\
\textbf{Q$_5$:} Is an enhancing capsule visible?\quad \textbf{A:} Yes.\\
\textbf{Q$_6$:} Is tumor in vein present?\quad \textbf{A:} No.
}
\end{algstepenv}

\begin{algstepenv}{\ding{193}~Step 2: Temporal Comparison (Foundation VQAs)}{steporange}{steporangebg}
{\small
\textbf{Q$_7$:} Was this lesion present in the prior scan?\quad \textbf{A:} No (\textsc{new}).\\
\textbf{Q$_8$:} How long since the prior imaging?\quad \textbf{A:} 5~months.
}
\end{algstepenv}

\begin{algstepenv}{\ding{194}~Step 3: Clinical Context (Foundation VQAs)}{stepgreen}{stepgreenbg}
{\small
\textbf{Q$_9$:} Cancer history?\quad \textbf{A:} Hepatocellular carcinoma (post-resection 2013).\\
\textbf{Q$_{10}$:} Is the patient at risk for HCC?\quad \textbf{A:} Yes.\\
\textbf{Q$_{11}$:} Is the LI-RADS pathway eligible?\quad \textbf{A:} Yes.
}
\end{algstepenv}

\begin{algstepenv}{\ding{195}~Step 4: Compose under LI-RADS v2018~\cite{chernyak2018liver}}{stepred}{stepredbg}
{\small
\textbf{Rule input.} At-risk (yes) $\land$ size $\geq$1\,cm (1.4\,cm) $\land$ arterial-phase hyperenhancement (yes) $\land$ washout (yes) $\land$ enhancing capsule (yes).\\
\textbf{Rule output.} \textbf{LR-5 (definite HCC).}\\[2pt]
\textit{Verification.} Pathology-confirmed HCC recurrence after 5-year remission.
}
\end{algstepenv}

\end{algbox}
\end{figure}

The chain trace in Box~1 is the central artifact \dataset\ releases. Every compositional VQA in the dataset comes with its trace recorded in this form. The atomic Q$_i$ at each step is itself a foundation VQA from \S\ref{sec:foundation_vqa}. The composition rule is the published rule of the governing standard. The final answer is verified by tissue diagnosis or by structured follow-up. We now define each tier formally.

\subsection{Foundation VQA: Atomic Perception}\label{sec:foundation_vqa}

Foundation VQAs ask questions whose answers are read directly from one field of the JSON record. They train atomic visual skills~\cite{hong2025atomicvisual,shao2024visualcot}. Examples grouped by source field follow.

\noindent\emph{Modality and acquisition.} ``What modality is this image?'' (CT). ``What is the contrast phase?'' (portal venous). ``What body region is shown?'' (abdomen). Source: scan metadata.

\noindent\emph{Anatomical presence.} ``Is the liver visible in this scan?'' (yes/no). ``Are the kidneys included?''. Source: canonical organ list and body-region tag.

\noindent\emph{Lesion presence and basic geometry.} ``Are any lesions present?''. ``How many lesions in the liver?''. ``What is the size of the largest hepatic lesion?''. ``Where is the lesion located?''. Source: \texttt{step1\_observations}.

\noindent\emph{Lesion attenuation and morphology.} ``Is the lesion hypodense in the portal-venous phase?''. ``Does the lesion have a calcified rim?''. Source: \texttt{step1\_observations}.

\noindent\emph{Patient demographics and history.} ``What is the patient's age?''. ``Does the patient have known cirrhosis?''. Source: \texttt{clinical\_variables} and \texttt{clinical\_history}.

\noindent\emph{Temporal change atoms.} ``Is the lesion present in both this scan and the prior?''. ``Did the longest diameter increase?''. Source: \texttt{step2\_temporal}.

\subsection{Single-Step Reasoning VQA}\label{sec:single_step_vqa}

Single-step reasoning VQAs apply one explicit clinical rule to one foundation observation. They require one inference step. Examples follow.

\noindent\emph{Threshold rules.} ``Is the renal lesion at or above 1\,cm by Bosniak criteria?''~\cite{silverman2019bosniak}. ``Is the liver lesion 2\,cm or larger by LI-RADS threshold?''~\cite{chernyak2018liver}. The atom is the size; the rule is the standard's threshold.

\noindent\emph{Single-feature classification.} ``Does the lesion show arterial-phase hyperenhancement?''. ``Does the lesion show portal-venous washout?''. The atom is the attenuation; the rule is the LI-RADS feature definition.

\noindent\emph{Single-change rules.} ``Is the lesion growing per RECIST 1.1?''~\cite{eisenhauer2009new}. The atom is the volume ratio; the rule is the 20\% threshold.

\noindent\emph{Single-context rules.} ``Is the patient eligible for the LI-RADS pathway?''. The atoms are cirrhosis status and chronic hepatitis B status; the rule is the LI-RADS at-risk definition.

\subsection{Compositional VQA: Multi-Step Chain-of-Thought}\label{sec:compositional_vqa}

Compositional VQAs require multiple foundation answers to be combined under the rules of a clinical reporting standard. The combination rule is given by the standard, not learned. The hard VQAs are the clinical decisions that radiologists actually make.

\noindent\emph{LI-RADS category.} ``What is the LI-RADS category of the largest hepatic lesion?''. Foundation chain: (i)~Is the patient at risk for HCC? (ii)~What is the lesion size? (iii)~Does it show arterial-phase hyperenhancement? (iv)~Does it show washout? (v)~Does it show an enhancing capsule? (vi)~Does it meet threshold growth? The composition rule maps these to LR-1 through LR-5.

\noindent\emph{RECIST response.} ``What is the RECIST 1.1 response category?''. Foundation chain: sum of target-lesion diameters at baseline and now, presence of new lesions, percent change. Rule: complete response, partial response, stable disease, or progressive disease.

\noindent\emph{TNM staging.} ``What is the T-stage of this gastric tumor?''~\cite{ajani2016gastric}. Foundation chain: wall thickness, serosal invasion, perigastric fat involvement, adjacent-organ invasion. Rule: TNM T1 to T4.

\noindent\emph{Differential under ambiguity.} ``Given imaging, history, and prior scans, what is the most likely diagnosis?''. Foundation chain: lesion characteristics, temporal change, clinical history, risk factors. Rule: the radiologist's standard differential workflow.

\subsection{Clinical Reporting Standards as Compositional Grammars}\label{sec:standards_grammars}

For every organ screening target, professional societies have codified a compositional grammar. LI-RADS~\cite{chernyak2018liver}, PI-RADS~\cite{turkbey2019prostate}, BI-RADS~\cite{d2018breast}, Bosniak~\cite{silverman2019bosniak}, and the others listed in Table~\ref{tab:standards} each define (a)~the atomic features to evaluate and (b)~a deterministic rule that maps these features to the final risk or diagnostic category. \dataset\ uses these grammars directly. The atomic features become foundation VQAs. The deterministic rule becomes the program for the compositional VQA. The four-step scaffold introduced next (\S\ref{sec:reasoning_traces}) is the canonical decomposition path: every compositional VQA is built from foundation VQAs that fall into one of four steps, namely imaging observations, temporal comparison, clinical context, and the diagnostic conclusion.

\section{Constructing Structured Clinical Reasoning Chains}\label{sec:reasoning_traces}

The four-step chain is the scaffold that decomposes compositional VQAs (\S\ref{sec:compositional_vqa}) into foundation VQAs (\S\ref{sec:foundation_vqa}). For each scan~$I_t$ with tumor mask $M_t$, radiology report $R_t$, clinical variables $C_t$, pathology $P$, and governing standard $\mathcal{S}$ (Table~\ref{tab:standards}), we construct
\begin{equation}\label{eq:trace}
\footnotesize
  \mathcal{T}_t \;=\; \bigl\langle\; \mathcal{O}_t,\;\; \Delta_t,\;\; \mathcal{C}_t,\;\; \mathcal{D}_t \;\bigr\rangle,
\end{equation}
where $\mathcal{O}_t$ are imaging observations grounded in $M_t$ and $\mathcal{S}$, $\Delta_t$ is the temporal change relative to prior scans, $\mathcal{C}_t$ is the clinical context parsed from $R_t$ and $C_t$, and $\mathcal{D}_t$ is the pathology-confirmed conclusion. We do not invent a reasoning vocabulary. We extract from each report the features that the governing standard prescribes. The construction pipeline (Algorithm~1 in Appendix~\ref{app:pipeline}) processes preprocessing, per-step extraction, and quality control. Per-step formal definitions and validation metrics are in Appendix~\ref{app:pipeline}.

\noindent\textbf{Validation.} The 200-patient cohort used for annotation QC (\S\ref{sec:cohort}) is reused for every pipeline step. Eight board-certified radiologists assess each step independently. The headline numbers are 62.2\% inter-annotator Dice for spatial annotations, 94.6\% feature-extraction accuracy (Step~1), 95.9\% temporal-label agreement (Step~2), 97.1\% report-parsing accuracy (Step~3), and Fleiss' $\kappa = 0.947$ for complexity stratification. Per-step breakdowns are in Appendix~\ref{app:pipeline}.

\noindent\textbf{Complexity stratification.} Each chain is labeled \textsc{perceptual}, \textsc{temporal}, \textsc{integrative}, or \textsc{ambiguous}, reflecting the information depth required to reach the conclusion. The decision rules and validation are in Appendix~\ref{app:pipeline}.

Table~\ref{tab:standards} maps each of the \numofclass\ organ screening targets to its governing standard. The features become foundation VQAs and the standard's rule becomes the program for compositional VQA.

\begin{table}[t]\label{sec:standards}
  \centering
  \caption{\textbf{Clinical reporting standards governing each cancer type in \dataset.} Each standard defines the imaging features, risk categories, and management recommendations used in clinical practice. Our reasoning chains extract report content and align it with these organ-specific feature vocabularies.}
  \label{tab:standards}
  \scriptsize
  \begin{tabular}{p{0.10\textwidth} p{0.19\textwidth} p{0.26\textwidth} p{0.33\textwidth}}
    \toprule
    \textbf{target organ} & \textbf{standard} & \textbf{key imaging features} & \textbf{risk stratification} \\
    \midrule
    thyroid & ACR IF \cite{hoang2015managing} & nodule size on CT, density, calcification, extrathyroidal extension & $<$1\,cm (ignore) to $>$2.5\,cm (further imaging) \\
    lung & Lung-RADS v2022 \cite{christensen2024acr} & nodule size, density (solid/GGO/part-solid), growth rate, spiculation & 1 (negative) to 4X (suspicious) \\
    breast & BI-RADS / ACR IF \cite{d2018breast, al2020practical} & mass density, margins, enhancement, calcification on CT & benign (ignore) to suspicious (tissue sampling) \\
    esophagus & NCCN + TNM \cite{ajani2019esophageal} & wall thickness, luminal narrowing, fat plane invasion & T/N staging criteria \\
    liver & LI-RADS \cite{chernyak2018liver} & arterial hyperenhancement, washout, enhancing capsule, threshold growth & LR-1 (benign) to LR-5 (definite HCC) \\
    gallbladder & ACR IF \cite{sebastian2013managing} & wall thickness, mucosal enhancement, polyp size & thin-wall (benign) to thick/enhancing (surgery) \\
    stomach & NCCN + Borrmann \cite{ajani2016gastric} & wall thickness, enhancement pattern, serosal invasion, morphology & Borrmann I--IV + T staging \\
    pancreas & ACR IF + Fukuoka \cite{megibow2017management,tanaka2017revisions} & duct dilation, mural nodules, solid component, cyst size & low/high-risk stigmata \\
    spleen & ACR IF \cite{heller2013managing} & lesion size, homogeneity, multiplicity, enhancement & $<$1\,cm (benign) to heterogeneous/growing (workup) \\
    duodenum & NCCN \cite{benson2019small} & mass size, obstruction, vascular encasement & resectability criteria \\
    colon & C-RADS \cite{pickhardt2003computed, zalis2005ct} & polyp size, morphology, location, number & C0--C4 categories \\
    kidney & Bosniak v2019 \cite{silverman2019bosniak} & septa, wall thickness, enhancement, calcification & I/II (benign) to IV (surgical) \\
    adrenal & ACR IF \cite{mayo2017management, herts2018management} & size, HU on unenhanced CT, washout characteristics & $\leq$1\,cm (ignore) to $>$4\,cm (surgery) \\
    bladder & VI-RADS \cite{panebianco2018multiparametric} & muscularis integrity, stalk morphology, signal on DWI & VI-RADS 1--5 \\
    prostate & PI-RADS v2.1 \cite{turkbey2019prostate} & T2 signal, DWI restriction, DCE, size, location by zone & PI-RADS 1--5 \\
    uterus & FIGO \cite{amant2018cancer} & endometrial thickness, myometrial invasion depth, cervical extension & FIGO stage I--IV \\
    ovary & O-RADS / ACR IF \cite{andreotti2020rads, atri2019acr} & cyst size, wall/septa thickness, solid component, enhancement & O-RADS 1 (normal) to 5 (high risk) \\
    lymph node & Lugano \cite{cheson2014recommendations} & short-axis diameter, morphology, enhancement, FDG avidity (PET) & measurable ($>$1.5\,cm) vs.\ non-measurable \\
    bone & WHO / RECIST \cite{who2020soft, eisenhauer2009new} & lytic/sclerotic morphology, cortical destruction, soft-tissue component & benign features to aggressive (biopsy) \\
    \bottomrule
  \end{tabular}
\end{table}

\section{Dataset Statistics and Analysis}\label{sec:statistics}

\begin{figure}[t]
  \centering
  \includegraphics[width=\linewidth]{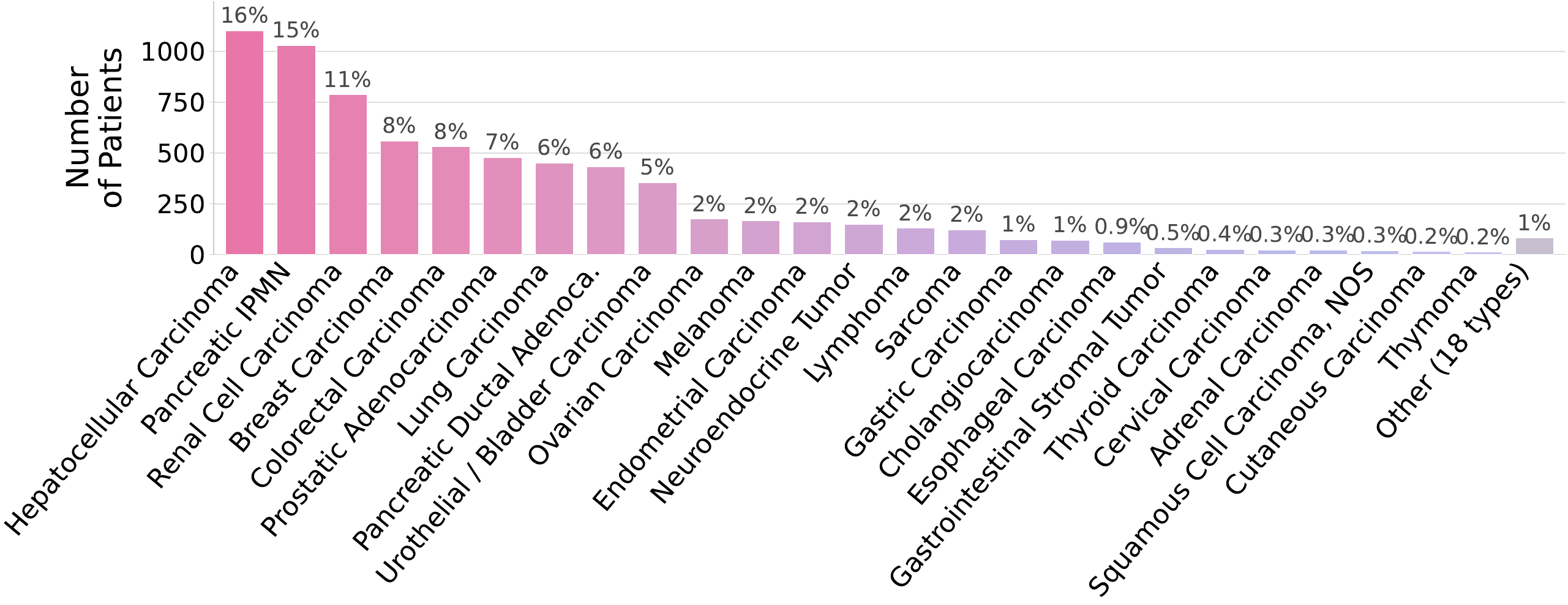}
  \caption{
    \textbf{Primary cancer type distribution across cancer-positive patients in \dataset.} Clinically equivalent subtypes are grouped together. For example, colon, rectal, and colorectal NOS are grouped into Colorectal Carcinoma. Bladder and urothelial cancers are grouped into a single category. The result is 43 distinct cancer groups with a pronounced long-tail distribution. Healthy controls (\numofnormal) are omitted for clarity.
  }
  \label{fig:cancer_dist}
\end{figure}

\dataset\ contains \numofct\ structured reasoning chains, one per CT scan, from \numofpatient\ patients.

\noindent\textbf{Cancer type distribution.} After merging clinically equivalent subtypes (Appendix~\ref{app:dedup}), the dataset spans 43 cancer groups (Fig.~\ref{fig:cancer_dist}) with a pronounced long-tail distribution. The five most frequent groups are hepatocellular carcinoma, pancreatic IPMN, renal cell carcinoma, breast carcinoma, and colorectal carcinoma. Together they account for over 50\% of cancer-positive patients. Of the 43 groups, 31 map to the \numofclass\ organ screening targets (Table~\ref{tab:standards}). The remaining 12 (507 patients) are cancers detected outside the screened organs via metastatic deposits or incidental findings (Appendix~\ref{app:cancer_organ_map}). Healthy controls account for \numofnormal\ patients (22.7\%). Among cancer-positive patients, 2{,}563 (36.3\%) have longitudinal imaging with 2 to 26 scans per patient.

\noindent\textbf{Reasoning complexity distribution.} Across all chains (Appendix~\ref{app:complexity}), \emph{integrative} reasoning has the largest share at 39.2\%. \emph{Ambiguous} follows at 36.4\%, then \emph{perceptual} at 12.9\%, and \emph{temporal} at 11.5\%. Integrative and ambiguous cases together cover 75.6\% of the dataset. This confirms that the majority of cancer screening requires reasoning beyond single-scan perception.

\noindent\textbf{Quality control summary.} Automated QC (Appendix~\ref{app:pipeline}) flagged 905 issues: timeline oscillations (304, concentrated in patients with $\geq$12 scans), uncertain primary cancer assignment (422), and malignancy flag inconsistencies (179).

\section{Training Vision-Language Models with \dataset}\label{sec:training}

\dataset\ supplies the supervised and RL stages of the modern VLM training stack. Modern open-source VLMs follow a four-stage recipe~\cite{wang2024qwen2,bai2025qwen25vl,dubey2024llama,steiner2024paligemma2,dai2024nvlm,lin2024vila,lu2024deepseekvl,wu2024deepseekvl2,li2024llavaonevision,zhu2025internvl3,agrawal2024pixtral,openai2024gpt4o}: unimodal pretraining, vision-language alignment, supervised fine-tuning (SFT), and preference or reinforcement-learning post-training. A second post-training axis is rule-based RL with verifiable rewards~\cite{guo2025deepseekr1,openai2024o1,comanici2025gemini25,microsoft2025phi4reasoning,huang2025visionr1,meng2025mmeureka,peng2025lmmr1,deng2025openvlthinker,liu2025visualrft}, mostly limited so far to math and perception~\cite{pan2025medvlmr1,lai2025medr1,jeong2025medprm,chen2024huatuogpt,wu2025medreason}. \dataset\ targets the SFT and RL stages with grounded multimodal CoT and pathology-confirmed rewards. We do not claim experimental results.

\noindent\textbf{Curriculum from foundation to compositional VQA.} The three tiers (\S\ref{sec:vqa_tiers}) form a natural SFT curriculum~\cite{bengio2009curriculum,deng2025currreft}. Foundation samples are (image, atomic question, short answer) triples that train visual skills~\cite{hong2025atomicvisual,shao2024visualcot}. Compositional samples are (image, hard question, four-step reasoning, answer) tuples that train chain-of-thought in the spirit of LLaVA-CoT~\cite{xu2024llavacot} and Visual-CoT~\cite{shao2024visualcot}. Box~2 shows examples for one patient. Two properties distinguish these chains from text-distilled CoT: they are grounded in voxel masks and clinical reporting standards, and their conclusions are verified by tissue diagnosis. The \numofct\ scans yield several hundred thousand SFT pairs across the three tiers.

\noindent\textbf{Verifiable rewards for RL.} The chain exposes four rule-based reward axes for GRPO-style RL~\cite{guo2025deepseekr1,pan2025medvlmr1,lai2025medr1}: pathology match against $\mathcal{D}_t$, organ-level malignancy and metastasis flags, organ-specific risk category $\kappa_{\mathcal{S}}$ with within-$\pm 1$ partial credit, and temporal change labels. A format reward enforces the four-step output structure. Details are in Appendix~\ref{app:rl}.

\noindent\textbf{Evaluation.} We recommend reporting accuracy stratified by VQA tier (foundation, single-step, compositional) and orthogonally by case complexity (\S\ref{app:complexity}). Together they show \emph{where} a model fails. Per-protocol details are in Appendix~\ref{app:eval}.

\noindent\textbf{Position against the open VLM ecosystem.} \dataset\ fills four gaps that frontier-lab reports leave open. Only Google publishes a medical VLM recipe (Med-Gemini~\cite{saab2024medgemini}, Med-PaLM~M~\cite{tu2024medpalmm}) and neither releases data. Most open VLMs use 2D backbones; \dataset\ provides 3D CT with voxel grounding. Reasoning-RL works repeatedly flag scarcity of multimodal CoT outside math~\cite{huang2025visionr1,meng2025mmeureka,peng2025lmmr1,deng2025openvlthinker,liu2025visualrft,microsoft2025phi4reasoning}; \dataset\ supplies grounded medical CoT. No other corpus supplies time-ordered scans with grounded change labels.

\begin{figure}[t]
\begin{algbox}{Box 2: VQA Examples Across the Three Tiers (Patient~P000001, Scan~2)}
\phantomsection\label{box:qa}%

\smallskip
{\footnotesize\textbf{Patient context.} Known metastatic breast carcinoma, post-mastectomy. Prior scan: 2017-07-24. Six lesions tracked across organs.}

\begin{algstepenv}{\ding{192}~Foundation VQA (atomic perception, no reasoning)}{stepblue}{stepbluebg}
{\small
\textbf{Q$_1$:} What modality is this image?\quad \textbf{A:} CT.\\
\textbf{Q$_2$:} Is the liver visible in this scan?\quad \textbf{A:} Yes.\\
\textbf{Q$_3$:} What is the size of the renal lesion in the left inferior pars?\quad \textbf{A:} 12\,mm.\\
\textbf{Q$_4$:} Was the right gluteal soft-tissue lesion present in the prior scan?\quad \textbf{A:} Yes.
}
\end{algstepenv}

\begin{algstepenv}{\ding{193}~Single-Step Reasoning VQA (one rule)}{steporange}{steporangebg}
{\small
\textbf{Q$_5$:} Is the renal lesion at or above 1\,cm by Bosniak criteria?\\
\textbf{A:} Yes (12\,mm $\geq$ 10\,mm threshold).\\[2pt]
\textbf{Q$_6$:} Has the right gluteal soft-tissue lesion resolved relative to the prior scan?\\
\textbf{A:} Yes (\textsc{resolved}; not present at the current time point).
}
\end{algstepenv}

\begin{algstepenv}{\ding{194}~Compositional VQA (multi-step chain-of-thought)}{stepred}{stepredbg}
{\small
\textbf{Q$_7$:} What is the RECIST 1.1 response category for this scan?\\
\textbf{A.}\\
\textbf{[Observations]} Five lesions present at both timepoints; renal lesion stable at 12\,mm.\\
\textbf{[Temporal]} One lesion \textsc{resolved}; no new lesions; no growth meeting RECIST progressive disease threshold.\\
\textbf{[Context]} Post-mastectomy follow-up; baseline 2018-01-24.\\
\textbf{[Conclusion]} Stable Disease per RECIST 1.1.\\[2pt]
\textbf{Q$_8$:} Synthesize imaging and clinical context to give the most likely diagnosis.\\
\textbf{A.} Metastatic breast carcinoma with stable disease. Pathology-confirmed.
}
\end{algstepenv}

\end{algbox}
\end{figure}

\section{Discussion and Conclusion}\label{sec:discussion_conclusion}

\dataset\ reframes cancer-screening AI as visual question answering. The three tiers form a curriculum from atomic perception to multi-step compositional reasoning. Every compositional VQA carries the chain of foundation VQAs that solves it. The chains are grounded in voxel-wise tumor masks and in organ-specific clinical reporting standards. The dominance of integrative and ambiguous cases (75.6\% of chains) confirms that most cancer screening lies beyond single-scan perception. The chain's quality is anchored in an eight-radiologist validation cohort: 62.2\% inter-annotator Dice, 94.6\% feature-extraction accuracy, 95.9\% temporal-label agreement, 97.1\% report-parsing accuracy, and Fleiss' $\kappa = 0.947$ for complexity stratification.

\noindent\textbf{Limitations.} \emph{(i) Information structure, not narrative.} Chains encode what was observed and how it changed. They do not encode free-form inferential narrative. \emph{(ii) Constructed from existing documentation.} The pipeline parses clinical reports rather than eliciting think-aloud protocols. \emph{(iii) CT only.} The four-step framework generalizes to mammography (BI-RADS) and MRI (PI-RADS) where standards exist. \emph{(iv) Incomplete multimodal data.} Some patients lack prior scans or structured reports; this mirrors clinical reality and provides an evaluation axis for reasoning under missing information. \emph{(v) Limited follow-up window.} The healthy cohort requires only $>$\nunofyearfollowup-year cancer-free follow-up. \emph{(vi) Point-estimate validation.} Quality metrics are reported without significance testing; downstream auditing~\cite{lubonja2025auditing} is encouraged.

\noindent\textbf{Conclusion.} \dataset\ supplies the medical VQA training data missing from current open VLM stacks: foundation-tier SFT for atomic visual skills, compositional-tier CoT SFT, and verifiable rewards for RL recipes such as DeepSeek-R1~\cite{guo2025deepseekr1,pan2025medvlmr1,lai2025medr1}. It complements parallel work on tumor synthesis~\cite{chen2024towards,yang2025medical}, continual learning for medical data~\cite{chou2024embracing,zhang2023continual}, and partial-label assembly~\cite{kang2023label}.

\begin{ack}
This work was supported by the Lustgarten Foundation for Pancreatic Cancer Research and the National Institutes of Health (NIH) under Award Number R01EB037669. We would like to thank the Johns Hopkins Research IT team in \href{https://researchit.jhu.edu/}{IT@JH} for their support and infrastructure resources where some of these analyses were conducted; especially \href{https://researchit.jhu.edu/research-hpc/}{DISCOVERY HPC}. We thank Yucheng Tang, Ho Hin Lee, Sucheng Ren, Junfei Xiao, Yuyin Zhou, and Jieneng Chen for their constructive suggestions at several stages of the project. We thank Jaimie Patterson for writing a news article about this project. Paper content is covered by patents pending.
\end{ack}

\clearpage
\bibliographystyle{abbrvnat}
\bibliography{refs,zzhou}


\newpage
\appendix

\section{JSON Schema of the Released Reasoning Chains}\label{app:json_schema}

This appendix documents the complete schema of the per-patient JSON file. Each patient is one JSON record. Each record has patient-level fields and a list of per-scan reasoning traces. We list every field with its type and value range. We then show a worked example for one scan.

\subsection{Patient-Level Fields}

\begin{table}[h]
  \centering
  \scriptsize
  \begin{tabular}{p{0.20\textwidth} p{0.10\textwidth} p{0.62\textwidth}}
    \toprule
    \textbf{field} & \textbf{type} & \textbf{description / value range} \\
    \midrule
    \texttt{patient\_id} & string & anonymized identifier (e.g., \texttt{P000001}) \\
    \texttt{primary\_cancer} & object & resolved primary cancer record (sub-fields below) \\
    \quad\texttt{primary\_cancer} & string & cancer type (e.g., \texttt{breast cancer}, \texttt{hepatocellular carcinoma}) \\
    \quad\texttt{confidence} & string & \texttt{high}, \texttt{medium}, or \texttt{low}; defined in App.~\ref{app:pipeline} \\
    \quad\texttt{source} & list[string] & resolver sources that voted (\texttt{icd10}, \texttt{tumor\_keyword}, \texttt{tumor\_flag}, \texttt{report\_nlp}, \texttt{report\_nlp\_definitive}) \\
    \quad\texttt{all\_candidates} & object & mapping from candidate name to weighted vote score \\
    \quad\texttt{metastasis\_sites} & list[string] & organs flagged as metastatic spread sites \\
    \quad\texttt{has\_metastatic\_disease} & bool & true if any metastasis site is flagged \\
    \texttt{clinical\_history} & list[string] & curated short statements about prior diagnoses, surgeries, oncological status \\
    \texttt{num\_scans} & int & number of CT scans in the longitudinal sequence \\
    \texttt{date\_range} & object & \texttt{first} and \texttt{last} ISO dates of the sequence \\
    \texttt{reasoning\_traces} & list[object] & one reasoning chain per scan (\S\ref{app:trace_fields}) \\
    \bottomrule
  \end{tabular}
\end{table}

\subsection{Per-Scan Trace Fields}\label{app:trace_fields}

Each entry in \texttt{reasoning\_traces} has the following five top-level fields plus the complexity label.

\paragraph{\texttt{metadata}.}
\texttt{patient\_id}, \texttt{scan\_id}, \texttt{accession} (string), \texttt{scan\_date} (ISO), \texttt{scan\_index} and \texttt{total\_scans} (int), \texttt{sex} and \texttt{age} (string, may be empty), \texttt{malignancy} and \texttt{metastasis} (string flags: \texttt{yes}, \texttt{no}, \texttt{u}).

\paragraph{\texttt{step1\_observations}.}
A list. Each finding has: \texttt{finding\_id} (string); \texttt{raw\_organ} and \texttt{canonical\_organ} (string, one of the \numofclass\ targets or an ancillary category); \texttt{location} and \texttt{standardized\_location} (string); \texttt{type} (free text describing tumor type); \texttt{type\_certainty} (\texttt{certain}, \texttt{high}, \texttt{low}); \texttt{size\_mm} (numeric, or \texttt{multiple}, or \texttt{U} for unknown); \texttt{attenuation} and \texttt{standardized\_attenuation} (string); \texttt{malignancy} and \texttt{metastasis} (\texttt{yes}, \texttt{no}, \texttt{u}); \texttt{clinical\_standard} (object with \texttt{name} and \texttt{reference}).

\paragraph{\texttt{step2\_temporal}.}
\texttt{status} is \texttt{no\_prior\_available} or \texttt{temporal\_comparison\_available}. When prior is available: \texttt{prior\_scan\_date} (ISO), \texttt{interval\_days} and \texttt{interval\_months} (numeric), \texttt{n\_matched}, \texttt{n\_new}, \texttt{n\_resolved} (int), and a list \texttt{changes}. Each change entry has: \texttt{organ}, \texttt{location}, \texttt{tumor\_type}, \texttt{matched\_with\_prior} (bool), and \texttt{change} (one of \texttt{NEW}, \texttt{GROWING}, \texttt{STABLE}, \texttt{SHRINKING}, \texttt{RESOLVED}, \texttt{PRESENT\_BOTH}). When sizes are available, \texttt{size\_current\_mm}, \texttt{size\_prior\_mm}, and \texttt{volume\_ratio} are populated; otherwise a \texttt{size\_note} explains the absence.

\paragraph{\texttt{step3\_clinical\_context}.}
\texttt{report\_parsed} is an object with \texttt{findings} (list of strings), \texttt{impression} (string), \texttt{recommendation} (string), and \texttt{parse\_method} (\texttt{rule\_based} or \texttt{llm}). \texttt{recist\_assessment} is one of \texttt{Stable Disease}, \texttt{Partial Response}, \texttt{Complete Response}, \texttt{Progressive Disease}, or null. \texttt{risk\_category} is the organ-specific category from $\kappa_{\mathcal{S}}$ (e.g., \texttt{LR-5}, \texttt{Bosniak IIF}, \texttt{PI-RADS 4}, \texttt{RECIST 1.1: Stable Disease}, or \texttt{not explicitly stated}). \texttt{clinical\_variables} stores \texttt{age}, \texttt{sex}, \texttt{contrast}, and \texttt{clinical\_history}. \texttt{raw\_report} is the de-identified original text.

\paragraph{\texttt{step4\_conclusion}.}
\texttt{primary\_cancer}, \texttt{primary\_cancer\_confidence}, \texttt{primary\_cancer\_source} (mirrors patient-level resolution at scan time). \texttt{has\_metastatic\_disease} (bool) and \texttt{metastasis\_sites} (list). \texttt{overall\_malignancy} and \texttt{overall\_metastasis} (\texttt{yes}, \texttt{no}, \texttt{u}). \texttt{icd10\_code} and \texttt{icd10\_organ} (string, may be empty). \texttt{organ\_level\_diagnosis} is an object mapping each organ that appears in this scan to a sub-object with \texttt{malignancy}, \texttt{metastasis}, and \texttt{primary\_tumor} fields.

\paragraph{\texttt{reasoning\_complexity}.}
One of \texttt{PERCEPTUAL}, \texttt{TEMPORAL}, \texttt{INTEGRATIVE}, \texttt{AMBIGUOUS} (\S\ref{app:complexity}).

\subsection{Worked Example}

The following abbreviated JSON shows Patient~P000001, Scan~2 (the same scan used in Box~2). Long lists are truncated for readability; the released file contains all findings.

{\scriptsize
\begin{verbatim}
{
  "patient_id": "P000001",
  "primary_cancer": {
    "primary_cancer": "breast cancer",
    "confidence": "high",
    "source": ["report_nlp", "report_nlp_definitive"],
    "has_metastatic_disease": true },
  "clinical_history": ["Known breast carcinoma", "Status post mastectomy", ...],
  "num_scans": 5,
  "date_range": {"first": "2017-07-24", "last": "2019-04-23"},
  "reasoning_traces": [
    {
      "metadata": {"scan_id": "S000002", "scan_date": "2017-11-28",
                    "scan_index": 2, ...},
      "step1_observations": [
        {"finding_id": "tumor 1", "canonical_organ": "chest_wall",
          "type": "metastasis", "malignancy": "yes",
          "clinical_standard": {"name": "RECIST 1.1 (general)"}}, ... ],
      "step2_temporal": {
        "status": "temporal_comparison_available",
        "interval_months": 4.2,
        "n_matched": 6, "n_new": 0, "n_resolved": 1,
        "changes": [
          {"organ": "kidney", "change": "STABLE", "volume_ratio": 1.0}, ...]},
      "step3_clinical_context": {
        "report_parsed": {
          "findings": [...],
          "impression": "Stable disease per RECIST 1.1...", ...},
        "recist_assessment": "Stable Disease",
        "risk_category": "RECIST 1.1: Stable Disease"},
      "step4_conclusion": {
        "primary_cancer": "breast cancer",
        "has_metastatic_disease": true,
        "organ_level_diagnosis": {
          "liver": {"malignancy": "yes", "metastasis": "yes"}, ...}},
      "reasoning_complexity": "INTEGRATIVE"
    }, ... ]
}
\end{verbatim}
}

\smallskip
The full released file preserves all findings, complete report text, and every clinical variable available at imaging time.


\newpage
\section{Reasoning Chain Construction Pipeline}\label{app:pipeline}

This appendix documents the formal definition and validation of each chain step (\S\ref{app:step1}--\S\ref{app:step4}), the construction algorithm (Algorithm~1, \S\ref{app:alg}), the complexity stratification rules (\S\ref{app:complexity}), and the data cleaning and quality control pipeline (\S\ref{app:cleaning}--\S\ref{app:qc}).

\subsection{Step 1: Imaging Observations}\label{app:step1}

From the voxel-wise tumor mask $M_t$, the CT volume $I_t$, and the governing clinical standard $\mathcal{S}$, we extract a structured observation for each finding. When $M_t \neq \emptyset$, the observation is a tuple
\begin{equation}\label{eq:obs}\footnotesize
  \mathcal{O}_t = \bigl(\, \mathrm{loc}(M_t),\;\; \mathrm{size}(M_t),\;\; \mathcal{F}_{\mathcal{S}}(R_t),\;\; \mathrm{morph}(I_t, M_t) \,\bigr).
\end{equation}
$\mathrm{loc}(\cdot)$ maps the mask centroid to an anatomical region via an organ atlas (e.g., ``liver, segment~VI''). $\mathrm{size}(\cdot)$ returns bounding-box axes and volume in cm$^3$ and flags standard-relevant thresholds. $\mathcal{F}_{\mathcal{S}}(R_t)$ are standard-specific descriptive features extracted from the findings section of $R_t$ via LLM-based parsing aligned to $\mathcal{S}$'s vocabulary. $\mathrm{morph}(\cdot)$ captures HU statistics, sphericity, surface irregularity, and calcification. When $M_t = \emptyset$, $\mathcal{O}_t$ records ``no suspicious findings.''

\noindent\textbf{Validation.} On the shared validation cohort ($n{=}200$), two independent reviewers per case yield 62.2\% inter-annotator Dice. For $\mathcal{F}_{\mathcal{S}}$, eight board-certified radiologists verified each extracted feature against the source report. Feature-level accuracy was 94.6\%. Errors concentrate in ambiguous modifiers (e.g., ``mildly heterogeneous'' vs.\ ``heterogeneous'').

\subsection{Step 2: Temporal Comparison}\label{app:step2}

For longitudinal scans, we co-register matched lesion pairs $(M_{t_{k-1}},\, M_{t_k})$ and compute the volume ratio $r = \mathrm{vol}(M_{t_k})\,/\,\mathrm{vol}(M_{t_{k-1}})$ to assign a temporal label:
\begin{equation}\label{eq:temporal}
\footnotesize
  \Delta_{t_k} =
  \begin{cases}
    \text{\textsc{new}}       & M_{t_{k-1}} = \emptyset,\; M_{t_k} \neq \emptyset \\
    \text{\textsc{growing}}   & r > 1.2 \\
    \text{\textsc{stable}}    & 0.8 \leq r \leq 1.2 \\
    \text{\textsc{shrinking}} & r < 0.8 \\
    \text{\textsc{resolved}}  & M_{t_{k-1}} \neq \emptyset,\; M_{t_k} = \emptyset
  \end{cases}
\end{equation}

The 20\% threshold follows RECIST-inspired volumetric criteria~\cite{eisenhauer2009new}. Three standards add their own temporal criteria. LI-RADS threshold growth is $\geq$50\% diameter increase in $\leq$6 months~\cite{chernyak2018liver}. Lung-RADS growth rate is a new solid component or $\geq$1.5\,mm mean diameter increase~\cite{christensen2024acr}. RECIST 1.1 progressive disease is $\geq$20\% sum-of-diameters increase~\cite{eisenhauer2009new}. When triggered, the chain carries both labels. For cancer-positive patients, the lead time $\ell_t = t_{\mathrm{index}} - t$ is the interval between each pre-diagnosis scan and pathological confirmation. Patients without prior scans receive $\Delta_t = \text{\textsc{no\_prior}}$.

\noindent\textbf{Validation.} On the validation cohort, eight radiologists reviewed all matched lesion pairs and their temporal labels. Agreement with automatic labels was 95.9\%. Lesion matching was correct in 193 of 200 cases (96.5\%). Errors involved small lesions ($<$1\,cm) in adjacent segments where co-registration ambiguity was unavoidable.

\subsection{Step 3: Clinical Context Integration}\label{app:step3}

Step~3 extracts the radiologist's interpretive synthesis (impression, risk stratification, recommendation) plus non-imaging clinical variables. The report $R_t$ and clinical variables $C_t$ are parsed into
\begin{equation}\label{eq:context}
\footnotesize
  \mathcal{C}_t = \bigl(\, F_t,\;\; \mathcal{I}_t,\;\; \mathcal{R}_t,\;\; \kappa_{\mathcal{S}}(\mathcal{I}_t),\;\; C_t \,\bigr),
\end{equation}
where $F_t$, $\mathcal{I}_t$, $\mathcal{R}_t$ are the extracted findings, impression, and recommendation, and $\kappa_{\mathcal{S}}(\cdot)$ maps the impression to the risk category defined by $\mathcal{S}$. Parsing uses rule-based section detection plus LLM extraction; non-standard reports ($\sim$12\%) use few-shot prompting. Risk categories are extracted in two paths. Regex-based extraction of explicitly stated categories (e.g., ``LR-4,'' ``Bosniak~IIF'') succeeds for 17.7\% of cancer scans. Rule-based derivation from observation features following each standard's published criteria fills the rest. Together these assign a risk category to 80.3\% of cancer scans. The remaining 19.7\% are governed by staging systems (FIGO, TNM) where risk-category scoring does not apply.

\noindent\textbf{Validation.} On the validation cohort, section-level parsing accuracy was 97.1\% (findings 98.9\%, impression 96.3\%, recommendation 98.2\%). Clinical variable linkage to scan timepoints passed 100\% temporal-integrity checks (no future-information leakage). Across all cancer scans, text-extracted and feature-derived risk categories agree exactly in 76.8\% of cases and within $\pm$1 category in 88.2\%.

\subsection{Step 4: Diagnostic Conclusion}\label{app:step4}

The final step anchors the chain to a definitive ground truth:
\begin{equation}\label{eq:conclusion}
\footnotesize
  \mathcal{D}_t =
  \begin{cases}
    (c,\, h) \;\text{from pathology } P & \text{if cancer-positive} \\
    \varnothing \;\text{($>$\nunofyearfollowup\text{-year follow-up})} & \text{if healthy}
  \end{cases}
\end{equation}
where $c$ is the cancer type and $h$ the histological subtype. Step~4 is a ground-truth anchor, not a reasoning step. It closes the chain so each becomes a self-contained evaluation unit.

\subsection{Algorithm: Reasoning Chain Construction}\label{app:alg}

\begin{figure}[h]
\begin{algbox}{Algorithm 1: Reasoning Chain Construction for a Single Scan}
\phantomsection\label{alg:trace}%

\smallskip
{\footnotesize\textbf{Input:} CT scan $I_t$; tumor mask $M_t$ ($\emptyset$ if healthy); prior scans $\{(I_{t_k}, M_{t_k})\}_{k<t}$; radiology report $R_t$; clinical variables $C_t$; pathology $P$; clinical standard $\mathcal{S}$}\\[1pt]
{\footnotesize\textbf{Output:} Structured reasoning chain $\mathcal{T}_t = \langle\, \mathcal{O}_t,\; \Delta_t,\; \mathcal{C}_t,\; \mathcal{D}_t \,\rangle$}

\begin{algstepenv}{\ding{192}~Step 1: Imaging Observations}{stepblue}{stepbluebg}
{\small
\textbf{if} $M_t \neq \emptyset$:\\
\quad Location $\leftarrow$ organ atlas on $M_t$;\;
Size $\leftarrow$ bbox axes + vol.\ (cm$^3$);\;
Morphology $\leftarrow$ sphericity, surface irregularity\\
\quad Standard features $\mathcal{F}_\mathcal{S} \leftarrow$ extract from $R_t$ aligned to $\mathcal{S}$\\
\quad $\mathcal{O}_t \leftarrow$ \textsc{StructuredObservation}(location, size, morphology, $\mathcal{F}_\mathcal{S}$)\\
\textbf{else}: $\mathcal{O}_t \leftarrow$ ``No suspicious findings''
}
\end{algstepenv}

\begin{algstepenv}{\ding{193}~Step 2: Temporal Comparison}{steporange}{steporangebg}
{\small
\textbf{if} prior $(I_{t_k}, M_{t_k})$ exists:\\
\quad Co-register $M_{t_k} \to M_t$;\; volume ratio $r = \text{vol}(M_t)/\text{vol}(M_{t_k})$\\
\quad $\Delta_t \leftarrow$ \textsc{new} / \textsc{growing} ($r{>}1.2$) / \textsc{stable} / \textsc{shrinking} ($r{<}0.8$) / \textsc{resolved}\\
\textbf{else}: $\Delta_t \leftarrow$ ``No prior available''
}
\end{algstepenv}

\begin{algstepenv}{\ding{194}~Step 3: Clinical Context Integration}{stepgreen}{stepgreenbg}
{\small
Parse $R_t \to (F_t,\; \mathcal{I}_t,\; \mathcal{R}_t)$ via rule-based detection + LLM\\
Extract risk category from $\mathcal{I}_t$ using $\mathcal{S}$ (LI-RADS, Bosniak, PI-RADS, \ldots);\;
clinical context $\leftarrow$ age, sex, history from $C_t$\\
$\mathcal{C}_t \leftarrow$ \textsc{ClinicalContext}($F_t,\; \mathcal{I}_t,\; \mathcal{R}_t,\; C_t$, risk category)
}
\end{algstepenv}

\begin{algstepenv}{\ding{195}~Step 4: Diagnostic Conclusion}{stepred}{stepredbg}
{\small
\textbf{if} cancer-positive: $\mathcal{D}_t \leftarrow$ cancer type and subtype from pathology $P$\\
\textbf{else}: $\mathcal{D}_t \leftarrow$ ``No malignancy, $>$1-year follow-up''
}
\end{algstepenv}

\end{algbox}
\end{figure}

\subsection{Reasoning Complexity Stratification}\label{app:complexity}

We stratify each chain into one of four complexity levels using rule-based decisions over dataset metadata.

\emph{Perceptual} cases satisfy all four conditions. (a)~Tumor longest axis $>$3\,cm. (b)~HU difference $>$20 between lesion and parenchyma. (c)~No ambiguity flag from annotation adjudication. (d)~High-risk category from imaging alone (LR-5, Bosniak~IV, PI-RADS~5). A single scan suffices.

\emph{Temporal} cases have a decisive temporal change label (\textsc{new}, \textsc{growing}, \textsc{resolved}) or meet a standard-specific temporal criterion such as LI-RADS threshold growth, but do not meet the Perceptual criteria. The finding becomes apparent only through longitudinal comparison.

\emph{Integrative} cases require synthesis of imaging with clinical context. The standard assigns an intermediate risk category (LR-3, Bosniak~IIF, PI-RADS~3). Reaching the conclusion demands integration of report impression, history, or demographics. There is no inter-radiologist discordance.

\emph{Ambiguous} cases carry the annotation protocol's ambiguity flag. Radiologists reached different clinical conclusions even after applying the standard. Three discordance types are tagged: \emph{boundary} (same diagnosis, different extent), \emph{classification} (different risk category), and \emph{detection} (presence vs.\ absence). Pathology is the definitive tiebreaker.

\noindent\textbf{Validation.} On the validation cohort, eight radiologists independently classified each patient. Inter-rater reliability was Fleiss' $\kappa = 0.947$. Agreement with automatic labels was 93\% (186/200). Of 14 disagreements, 9 were Integrative$\leftrightarrow$Ambiguous, 3 Perceptual$\leftrightarrow$Temporal, and 2 other.

\subsection{Data Cleaning}\label{app:cleaning}

Constructing chains at scale requires four cleaning operations. (1)~\emph{Organ name normalization} maps 191 raw surface forms to \numofclass\ canonical targets. (2)~\emph{Primary cancer resolution} combines ICD-10 codes, tumor-type keywords, malignancy/metastasis flags, and report NLP through weighted majority voting. (3)~\emph{Lesion-level temporal tracking} matches findings across scans by organ, location, and type, with fuzzy organ-family grouping. (4)~\emph{Clinical validation} ensures that metastasis sites are never confused with primary cancers. Pipeline architecture and quality control flag statistics follow.

\subsection{Pipeline Architecture}

The pipeline takes two data sources as input. (1)~Per-scan metadata. This includes radiology reports, ICD-10 codes, clinical variables, and RECIST assessments (\numofct\ rows). (2)~Per-tumor metadata. This includes organ labels, tumor types, locations, sizes, and malignancy flags (45{,}641 rows). The pipeline produces structured 4-step reasoning chains following Algorithm~1.

\paragraph{Organ name normalization.}
Raw organ labels in the source data contain 191 unique surface forms. Examples include ``chest wall,'' ``thoracic wall,'' and ``pectoral region.'' We must map these to the \numofclass\ canonical cancer screening targets plus categorized ancillary organs. We built a two-pass normalization table. The first pass uses exact string matching. The second pass uses substring matching. Together they consolidate all 191 forms into canonical names. We also define organ \emph{families} (e.g., breast $\approx$ chest wall $\approx$ soft tissue) for fuzzy temporal matching.

\paragraph{Multi-source primary cancer resolver.}
Determining each patient's primary cancer is essential for grounding the clinical context step. It is non-trivial. ICD-10 codes are available for only 27\% of scans. Tumor type labels are heterogeneous (722 unique strings). Radiology reports describe findings without always stating the diagnosis explicitly. Our resolver combines four sources with weighted voting. (1)~\emph{ICD-10 primary tumor codes} (weight~4) use only the C00--C76 range for primary neoplasms. We record metastasis codes (C77--C79) solely as spread sites. This prevents clinically incorrect labels such as ``bone metastasis'' being assigned as a primary cancer. (2)~\emph{Tumor type keywords} (weight~3) provide a curated mapping of 180+ tumor type strings to canonical primary cancer names. (3)~\emph{Tumor-level flags} (weight~2) infer the primary cancer from the organ of origin for each tumor with \texttt{malignancy=yes} and \texttt{metastasis=no}. (4)~\emph{Two-tier report NLP} treats definitive radiologist statements such as ``metastatic breast cancer'' or ``known melanoma'' with weight~5. It treats 50+ standard regex patterns for cancer names, surgical procedures, and diagnostic signs with weight~1 to 3.
Votes are aggregated at the patient level across all scans. The candidate with the highest weighted score is selected. Confidence is assigned as \emph{high}, \emph{medium}, or \emph{low}. \emph{High} requires the dominant candidate to score $\geq 2\times$ runner-up with total $\geq$3. \emph{Medium} requires the dominant candidate to exceed the runner-up. \emph{Low} covers the cases with no candidate or a tie.

\paragraph{Lesion-level temporal tracking.}
For patients with longitudinal imaging, we match lesions across consecutive scans using a two-pass algorithm. (1)~Exact match on (canonical organ, location, tumor type). (2)~Fuzzy match using organ family grouping for unmatched lesions. Matched lesions receive temporal change labels based on size ratios. The labels are growing ($>$1.2$\times$), shrinking ($<$0.8$\times$), or stable. Unmatched prior lesions are labeled \emph{resolved}. Unmatched current lesions are labeled \emph{new}.

\paragraph{Two-stage report parsing.}
Radiology reports are parsed in two stages. (1)~Rule-based section detection identifies headers and dash-delimited lists. It extracts structured findings and RECIST assessments. (2)~A sentence-splitting fallback handles unstructured narrative text. Recommendations and clinical impressions are separated from imaging findings.

\subsection{Quality Control Flags}\label{app:qc}

Automated QC checks are applied to every patient. They flag three categories of issues. \emph{Timeline oscillations} (304 flags) occur when a lesion appears resolved and then reappears in a later scan. These flags concentrate in patients with $\geq$12 scans. They reflect radiologist reporting variability rather than true biological change. \emph{Primary cancer uncertain} flags (422) arise when primary cancer confidence remains low despite all four resolver sources being consulted. They typically appear in patients whose reports describe non-specific findings without clearly indicating a primary malignancy. \emph{Malignancy flag inconsistencies} (179 flags) occur when the malignancy label for a lesion changes across scans without an intervening treatment or biopsy event.


\newpage
\section{Training-Path Details}\label{app:training_details}

This appendix expands on the SFT, RL, and evaluation paths summarized in Section~\ref{sec:training}.

\subsection{Verifiable Rewards for Reinforcement Learning}\label{app:rl}

The four-step structure exposes four reward signals that are deterministic and require no further human annotation. They fit GRPO-style RL recipes used by DeepSeek-R1~\cite{guo2025deepseekr1} and its medical variants~\cite{pan2025medvlmr1,lai2025medr1}.

\emph{(i) Pathology match.} The model's predicted cancer type is compared to $\mathcal{D}_t$. Exact match yields reward 1; mismatch yields 0. Available for every cancer-positive patient.

\emph{(ii) Organ-level malignancy and metastasis.} For each organ in \texttt{organ\_level\_diagnosis}, the predicted (malignancy, metastasis) flags are checked against ground truth. The reward decomposes into a sum across organs and is multi-label.

\emph{(iii) Risk category.} The predicted LI-RADS, PI-RADS, Bosniak, or analogous category is compared to $\kappa_{\mathcal{S}}(\mathcal{I}_t)$. Exact match is rewarded; within-$\pm 1$ category counts as partial credit.

\emph{(iv) Temporal change.} For multi-scan inputs, the predicted change label per lesion is checked against $\Delta_{t_k}$.

A format reward enforces the four-step output structure. To our knowledge, \dataset\ is the first public resource that provides all four signals for cancer screening at scale.

\subsection{Recommended Evaluation Protocols}\label{app:eval}

We recommend reporting accuracy at each VQA tier, with case complexity (\S\ref{app:complexity}) as an orthogonal axis.

\noindent\textbf{Foundation accuracy.} Closed-form atomic perception questions (modality, contrast phase, organ presence, lesion presence, lesion size, attenuation). Matches the closed-question protocol of VQA-RAD~\cite{lau2018dataset}, SLAKE~\cite{liu2021slake}, and OmniMedVQA~\cite{hu2024omnimedvqa}.

\noindent\textbf{Single-step reasoning accuracy.} One clinical rule applied to one foundation observation. Threshold checks, single-feature classification, and single-change rules. Reports whether the model knows the rule but fails the perception, or vice versa.

\noindent\textbf{Compositional accuracy.} Multi-step questions ending in a clinical-guideline category (LI-RADS, PI-RADS, Bosniak, RECIST, TNM). Pathology serves as the terminal ground truth where applicable. Matches the medical VQA setup of LLaVA-Med~\cite{li2024llava} and Med-R1~\cite{lai2025medr1}, but with the chain of foundation answers exposed for step-level diagnosis.

\noindent\textbf{Longitudinal compositional accuracy.} Multi-scan input. The model must integrate the temporal trajectory before answering. The longitudinal cohort in \dataset\ (2{,}563 cancer patients with 2 to 26 scans) is sized for this protocol.

\subsection{Integration Recipe}\label{app:integration}

A team integrating \dataset\ into an existing VLM stack mixes the foundation and compositional SFT pairs into Stage~3 (a 5 to 15\% mix ratio avoids oncology overfitting), uses the four verifiable rewards plus a four-step format reward in a Stage~4 GRPO pass, and reports results stratified by VQA tier and case complexity. Aggregate accuracy hides where the gains come from. Stratification keeps temporal and integrative gains visible.


\newpage
\section{Illustrative Patient: Scan-by-Scan Reasoning Annotations}\label{app:illustrative}

The following provides the full reasoning chain annotations for the six selected timepoints of the illustrative HCC patient shown in Figure~\ref{fig:dataset_overview}. This patient was monitored over 26~CT scans across 11~years (2013--2024).

\smallskip
\textbf{Scan~1 (2013-07):} \textit{Observation:} hemangioma in liver segment~VII; post-resection changes in segment~VIII. \textit{Temporal:} no prior. \textit{Context:} report impression: ``status post resection of segment~VIII without evidence of recurrence.'' \textit{Complexity:} \textsc{Integrative}. The hemangioma must be distinguished from recurrent HCC using clinical context, not imaging alone.

\smallskip
\textbf{Scan~2 (2013-10):} \textit{Temporal:} \textbf{new} ill-defined hypervascular area near resection margin. \textit{Context:} ``likely perfusion alteration, but HCC recurrence cannot be excluded; follow-up in 3~months.'' \textit{Complexity:} \textsc{Temporal}. The new lesion is the decisive finding. Its nature is still ambiguous.

\smallskip
\textbf{Scan~5 (2014-08):} \textit{Temporal:} prior suspicious lesion \textbf{resolved}; hemangioma stable. \textit{Context:} ``complete remission; stable postoperative cyst.'' \textit{Complexity:} \textsc{Temporal}. The resolution event is the key reasoning step. It confirms that the prior finding was benign.

\smallskip
\textbf{Scan~17 (2019-12):} \textit{Observation:} \textbf{new} hypervascular lesion in segments~VI/VII. \textit{Context:} ``LI-RADS~5, suspicious for HCC recurrence;'' also a LI-RADS~3 lesion (indeterminate). \textit{Complexity:} \textsc{Temporal}. A new lesion after 5 years of remission changes the clinical trajectory.

\smallskip
\textbf{Scan~24 (2023-03):} \textit{Observation:} three lesions (13\,mm, 6\,mm, 3\,mm) in segments~V/VII after microwave ablation. \textit{Temporal:} one \textbf{growing}, two \textbf{new}. \textit{Context:} ``multifocal HCC; LI-RADS~3 lesions suspicious for recurrence; recommend liver MRI.'' \textit{Complexity:} \textsc{Temporal}. The pattern is recurrence after ablation.

\smallskip
\textbf{Scan~26 (2024-04):} \textit{Observation:} two LI-RADS~3 lesions, one with arterial enhancement. \textit{Context:} ``no new hepatic lesions; known LI-RADS~3 lesions without progression.'' \textit{Complexity:} \textsc{Integrative}. Stability over 8 months, combined with LI-RADS criteria, downgrades concern. \textit{Conclusion:} HCC, pathology-confirmed.


\newpage
\section{Data Normalization Vocabulary}\label{app:dedup}

The raw dataset contains heterogeneous labels from pathology records, radiology reports, and ICD-10 codes. This appendix documents the complete normalization vocabulary used to construct structured reasoning chains. All original labels are preserved in the released data alongside canonical forms.

\paragraph{Organ name normalization.}
Table~\ref{tab:organ_vocab} maps all 191 raw organ surface forms to the \numofclass\ canonical screening targets plus ancillary organ categories. The normalization table was built by two-pass matching: exact string matching followed by substring matching.

\begin{table}[h]
  \centering
  \caption{\textbf{Organ name normalization vocabulary for the \numofclass\ organ screening targets.} Each canonical organ is shown with its raw surface forms from the source data (observation count in parentheses). The 45 non-target organ categories (64 additional surface forms) follow RECIST~1.1~\cite{eisenhauer2009new} general criteria.}
  \label{tab:organ_vocab}
  \scriptsize
  \begin{tabular}{p{0.10\textwidth} p{0.04\textwidth} p{0.78\textwidth}}
    \toprule
    \textbf{canonical} & \textbf{$N$} & \textbf{raw surface forms (observation count)} \\
    \midrule
    thyroid & 2 & thyroid (255), thyroid gland (3) \\
    lung & 2 & lung (5{,}343), lungs (6) \\
    breast & 1 & breast (319) \\
    esophagus & 2 & esophagus (98), esophagus/stomach (1) \\
    liver & 3 & liver (13{,}034), lung/liver (1), liver/kidney (1) \\
    gallbladder & 1 & gallbladder (77) \\
    stomach & 2 & stomach (185), stomach/small intestine (1) \\
    pancreas & 2 & pancreas (3{,}035), duodenum/pancreas (1) \\
    spleen & 2 & spleen (748), liver/spleen (1) \\
    duodenum & 1 & duodenum (80) \\
    colon & 8 & colon (684), appendix (16), rectum (12), anal canal (1), anal region (1), small intestines/colon (1), stomach/colon (1), colon/duodenum (1) \\
    kidney & 5 & kidney (3{,}991), left kidney (78), right kidney (48), kidneys (3), renal fossa (1) \\
    adrenal & 6 & adrenal gland (1{,}501), adrenal (18), adrenal glands (7), adrenal gland/kidney (1), right adrenal gland (1), adrenal gland/bone (1) \\
    bladder & 2 & bladder (296), urinary bladder (6) \\
    prostate & 1 & prostate (170) \\
    uterus & 3 & uterus (277), cervix (6), uterus/vagina (1) \\
    ovary & 4 & ovary (288), ovaries (2), right ovary (1), left ovary (1) \\
    lymph\_node & 14 & lymph nodes (1{,}002), lymph node (782), axilla (19), supraclavicular (3), axillary (3), inguinal canal (2), cervical (2), cervical/supraclavicular (1), mediastinal lymph nodes (1), mediastinal (1), retroperitoneal lymph nodes (1), lymphatic system (1), interaortocaval (1), infracarinal (1) \\
    bone & 16 & bone (3{,}957), spine (115), axial skeleton (8), rib (7), skeleton (6), sacrum (4), femur (4), skeletal system (3), skull (2), vertebra (2), skeletal (2), ilium (1), sternum (1), rib cage (1), scapula (1), paravertebral (1) \\
    \midrule
    \multicolumn{2}{l}{\textit{+ 45 non-target}} & peritoneum (5 forms), pelvis, soft\_tissue (3), muscle (7), skin (3), chest\_wall (4), vascular (18), mediastinum, pleura, small\_intestine (5), retroperitoneum (2), abdomen, abdominal\_wall, neck (2), thorax (2), \ldots \\
    \bottomrule
  \end{tabular}
\end{table}

\paragraph{Clinical deduplication of cancer types.}
Table~\ref{tab:dedup} documents all seven merges of clinically equivalent cancer subtypes into unified groups, reducing the label set from 55 to 43 clinically distinct cancer groups.

\begin{table}[h]
  \centering
  \caption{\textbf{Clinical deduplication of cancer type labels.} Seven groups of clinically equivalent subtypes are merged. All original fine-grained labels are preserved in the released data.}
  \label{tab:dedup}
  \scriptsize
  \begin{tabular}{p{0.22\textwidth} p{0.38\textwidth} p{0.32\textwidth}}
    \toprule
    \textbf{unified group} & \textbf{original labels merged} & \textbf{clinical justification} \\
    \midrule
    colorectal carcinoma & colon cancer (347), rectal cancer (100), colorectal cancer NOS (80), rectosigmoid cancer (3) & anatomically contiguous segments of the large bowel; shared TNM staging, NCCN guidelines, and screening protocols \\
    lung carcinoma & lung cancer (430), small cell lung cancer (14), non-small cell lung cancer (7) & all primary lung malignancies; subtypes share the same organ screening target and imaging vocabulary \\
    pancreatic ductal adenocarcinoma & pancreatic cancer NOS (280), pancreatic ductal adenocarcinoma (152) & PDAC accounts for $>$85\% of pancreatic cancers; NOS labels typically reflect unspecified histology rather than a distinct entity \\
    urothelial / bladder ca. & bladder cancer (216), urothelial carcinoma (137) & urothelial carcinoma constitutes $>$90\% of bladder malignancies; both labels refer to the same disease \\
    endometrial carcinoma & endometrial cancer (121), uterine cancer (39) & endometrial carcinoma is the dominant uterine malignancy ($>$90\%); ``uterine cancer'' without qualifier denotes endometrial origin \\
    neuroendocrine tumor & neuroendocrine tumor (133), pancreatic NET (12), neuroendocrine carcinoma (3) & shared neuroendocrine lineage; grouped for statistical power while acknowledging grade heterogeneity (NET G1/G2 vs.\ NEC G3) \\
    cholangiocarcinoma & cholangiocarcinoma NOS (55), intrahepatic CCA (10), extrahepatic CCA (6) & subtypes of the same biliary epithelial malignancy; anatomic subsite preserved in released chain metadata \\
    \bottomrule
  \end{tabular}
\end{table}

\newpage
\section{Cancer Group to Organ Screening Target Mapping}\label{app:cancer_organ_map}

Each reasoning chain is governed by the clinical reporting standard of the organ in which a finding is detected (Table~\ref{tab:standards}). The patient's \emph{primary cancer} may differ from the screened organ. For example, a breast cancer patient may present with liver metastases evaluated under LI-RADS. Table~\ref{tab:cancer_organ} maps all 43 cancer groups to the \numofclass\ organ screening targets. Groups marked \emph{extra-organ} represent primary cancers outside the screened organs, detected via metastatic deposits or incidental findings on CT.

\begin{table}[h]
  \centering
  \caption{\textbf{Mapping of 43 cancer groups to the \numofclass\ organ screening targets.} 31 groups map to a screened organ. 12 are ``extra-organ'' primary cancers detected via metastatic deposits or incidental findings on CT.}
  \label{tab:cancer_organ}
  \scriptsize
  \begin{tabular}{p{0.14\textwidth} p{0.52\textwidth} r}
    \toprule
    \textbf{organ target} & \textbf{cancer groups} & \textbf{patients} \\
    \midrule
    thyroid & thyroid carcinoma & 26 \\
    lung & lung carcinoma & 451 \\
    breast & breast carcinoma & 559 \\
    esophagus & esophageal carcinoma & 61 \\
    liver & hepatocellular carcinoma, cholangiocarcinoma, hepatoblastoma & 1{,}172 \\
    gallbladder & gallbladder carcinoma & 12 \\
    stomach & gastric carcinoma, gastrointestinal stromal tumor & 107 \\
    pancreas & pancreatic IPMN, pancreatic ductal adenoca., pancreatobiliary ca., pancreatic neoplasm (benign) & 1{,}466 \\
    spleen & splenic neoplasm & 9 \\
    duodenum & duodenal carcinoma, ampullary carcinoma & 13 \\
    colon & colorectal carcinoma, anal carcinoma & 532 \\
    kidney & renal cell carcinoma, clear cell carcinoma & 787 \\
    adrenal & adrenal carcinoma & 22 \\
    bladder & urothelial / bladder carcinoma & 353 \\
    prostate & prostatic adenocarcinoma & 477 \\
    uterus & endometrial carcinoma, cervical carcinoma, vulvar carcinoma & 184 \\
    ovary & ovarian carcinoma, ovarian borderline tumor & 177 \\
    lymph node & lymphoma & 129 \\
    bone & multiple myeloma & 10 \\
    \midrule
    \multicolumn{2}{l}{\textit{subtotal: organ-mapped (31 groups)}} & \textit{6{,}547} \\
    \midrule
    \multirow{3}{*}{\parbox{0.14\textwidth}{extra-organ\\(detected via\\metastasis or\\incidental finding)}}
      & melanoma, neuroendocrine tumor, sarcoma, squamous cell carcinoma (NOS), & \\
      & cutaneous carcinoma, thymoma, mesothelioma, testicular carcinoma, & \\
      & laryngeal ca., pharyngeal ca., oropharyngeal ca., small intestine ca. & 507 \\
    \midrule
    \multicolumn{2}{l}{\textit{subtotal: extra-organ (12 groups)}} & \textit{507} \\
    \midrule
    \multicolumn{2}{l}{\textbf{total cancer-positive patients}} & \textbf{7{,}054} \\
    \bottomrule
  \end{tabular}
\end{table}


\end{document}